\documentclass{article}
\usepackage{iclr2027_conference,times}
\usepackage{amsmath,amssymb}
\usepackage{booktabs}
\usepackage{graphicx}
\usepackage{multirow}
\usepackage{xcolor}
\usepackage{hyperref}
\hypersetup{hidelinks,pdftitle={Tracking Is Not Permanence: What Video World Models Keep of a Hidden Object},pdfauthor={Peng Xie, Amr Alanwar}}
\usepackage{url}
\newcommand{\pres}{P(\text{present})}
\newcommand{\best}[1]{\textbf{#1}}

\title{Tracking Is Not Permanence:\\What Video World Models Keep of a Hidden Object}

\author{Peng Xie\thanks{Correspondence to \texttt{p.xie@tum.de}.} \ \& Amr Alanwar \\
TUM School of Computation, Information and Technology \\
Technical University of Munich, Germany}

\iclrfinalcopy

\begin{document}
\maketitle
\lhead{Preprint. Under review.}

\begin{abstract}
Video world models track objects they can see; we ask what they keep of objects they cannot. We hide an object from a frozen V-JEPA~2 predictor and compare its prediction for the hidden region with the encoder's representation of two worlds that differ only inside that region. Its decision keeps a stationary object in part and one carried inside a container not at all, and loses a moving one within 0.3\,s, or 0.5\,s under V-JEPA's own tube mask (ViT-H keeps it to 1.1\,s at pretraining's 90\% masking ratio); in projection a trace remains after the first tubelet, below the midpoint for a moving object and a closed box, at 14--60\% of what a baseline copying the last view retains. The information is there: the encoder reads the object's presence at 1.00 and keeps a closed container's contents decodable for the 3.5\,s after the event, while the predictor's output, read with the encoder's own probe, contains the ball in 2\% of scenes once the box has been closed for half a second. On rendered scenes permanence is missing on the predictor's side, and training installs it cheaply, as a prior: three thousand predictor-only steps on synthetic containers take this belief from 0.05 to 1.00 against two matched controls. They also raise IntPhys-2019 from 84.2\% to 93.3\%, but so does a curriculum without containers, and which training habit the benchmark credits changes with its scoring rule: for these checkpoints, benchmark accuracy does not track the belief. Continued training with tube masks for 16k steps produces 1.1--1.6\,s of moving-object carry-over on manipulation and internet-style video, so the deficit is not intrinsic to latent prediction; what persists changes with the masks and the data. VideoMAE's pixel reconstruction keeps almost nothing, and Cosmos's next-token prediction keeps a stationary hidden object but not one carried inside a moving container, where V-JEPA~2 and Cosmos fail alike.
\end{abstract}

\section{Introduction}

A world model that plans must keep state for what is no longer in view. A hand that closed over a coin still holds it. Infants show this expectation at five months \citep{baillargeon1985object}, and it is separable from tracking a visible object \citep{spelke1992origins,piaget1954construction}. Self-supervised video models are now used as world models for prediction and planning \citep{assran2025vjepa2,nvidia2025cosmos} and score above chance on intuitive-physics benchmarks built around occlusion \citep{garrido2025intuitive}, yet whether they keep any state for an occluded object has not been read from the predictor's own state (concurrent work with other readouts: Appendix~\ref{app:scope}).

We started from filmed impossible events, where an object leaves a closed hand: across three datasets and six readouts, V-JEPA~2's prediction error tracks motion energy and never the human-rated moment of surprise (Appendix~\ref{app:magic}); telling a missing expectation from a blind readout takes an instrument that reads the state itself.

The instrument is a belief probe. We render minimal pairs in which an object is hidden and the two futures differ only inside the hidden region, give the frozen encoder the visible context, let the frozen predictor infer the hidden tokens, and compare the prediction with the encoder's own representation of each future. The readout is a two-alternative choice with chance at 0.5 and a bias control, an object that was never shown; it applies unchanged to a pixel-reconstruction model and in a teacher-forced form to an autoregressive one. The model under test is the released V-JEPA~2 pretraining predictor at three scales, frozen, on rendered scenes; the action-conditioned V-JEPA~2-AC and V-JEPA~2.1 are not probed, and Appendix~\ref{app:scope} bounds every claim.

The probe gives a sharp answer: under the main query (band hidden from the event on, frame otherwise visible) V-JEPA~2 keeps a just-hidden object for one step and then stops, $\pres$ falling below chance between the first and second tubelet (a two-frame slot, 0.27\,s) after occlusion at all three scales (0.92 to 0.00 for ViT-H), while the same encoder reads the object's presence at 1.00 and keeps a rigid box's contents decodable for the 3.5\,s after the event. That gap is the paper's central measurement; we then show what does and does not produce permanence when we train, and where pixel-reconstruction and autoregressive models fail.

Our contributions:
\begin{itemize}
\item A belief probe, a per-tubelet violation-of-expectation test on the hidden tokens, that reads permanence from a frozen predictor's latents against matched controls (Section~\ref{sec:method}).
\item In the released V-JEPA~2 a moving object reads as \emph{absent} within 0.3\,s at all three scales under the main query and within 0.5\,s under V-JEPA's own tube geometry (at pretraining's 90\% masking ratio ViT-L and ViT-g still lose it within 0.3\,s; ViT-H keeps it to 1.1\,s); a stationary one is kept in 13--41\% of scenes on average (control at most 7\%, a copy of the last view 97--98\%), and revealing the object at the clip's end moves only the adjacent tubelet: the predictor fills in locally (Sections~\ref{sec:curves}--\ref{sec:reveal}).
\item On rendered scenes the deficit is on the predictor's side: presence is linearly decodable from the frozen encoder at 1.00 and a box's contents at 0.91--1.00 for the 3.5\,s after the event, while the encoder's own probe finds the ball in the predictor's output for the same region in 2\% of scenes at 0.53\,s and in none afterwards (Section~\ref{sec:encoder}).
\item The deficit is not intrinsic to latent prediction: each mask kind we tried on synthetic scenes yields 1.3--1.6\,s of carry-over, continued training with tube masks for 16k steps produces it on egocentric and internet-style video (on natural video a stationary object is kept only by mixtures that include temporally-partial masks), and adding temporally-partial masks makes the model freeze the object instead (Section~\ref{sec:objective}).
\item VideoMAE keeps almost nothing, and the autoregressive Cosmos-Predict1-4B keeps a stationary hidden object for over 2\,s but not one carried inside a moving container, where V-JEPA~2 and Cosmos fail alike (Section~\ref{sec:paradigms}).
\item Three thousand predictor-only steps on synthetic containers install a containment prior (1.00 against two controls at 0.00) and raise IntPhys-2019 from 84.2\% to 93.3\% for ViT-H (61.4\% to 64.4\% for ViT-L, 54.2\% to 75.0\% for ViT-g) with two context frames, and under every other rule we tried, but control curricula show that which habit earns the gain depends on the rule, so, for these checkpoints, the benchmark does not track the belief. Training the encoder on real hands separates coin-in, removed and never on held-out clips and on two unseen participants (Section~\ref{sec:install}).
\end{itemize}

\section{Related Work}

\paragraph{Intuitive physics benchmarks.} IntPhys \citep{riochet2022intphys} and IntPhys~2 \citep{bordes2025intphys2} pair possible and impossible videos and score a model by whether its prediction error is higher for the impossible one. \citet{bordes2025intphys2} report 87.2\% for V-JEPA~2 ViT-H and 85.4\% for Cosmos-Predict1-4B on IntPhys-2019 with this readout, each the best of several runs; we obtain 84.2\% under the fixed-context variant of \citet{garrido2025intuitive}, whose appendix raises, without testing it, the question of Section~\ref{sec:curves}. Physion \citep{bear2021physion}, InfLevel \citep{weihs2022benchmarking} and the minimal video pairs of \citet{krojer2025mvp} score contact prediction, infant-style contrasts and shortcut-controlled question answering. Closest to us are LA-CATER \citep{shamsian2020learning}, built on CATER \citep{girdhar2020cater}, and TCOW \citep{vanhoorick2023tcow}, which tracks objects through containers and occluders in synthetic and real video and finds video transformers far from permanence; the Perception Test \citep{patraucean2023perception} films a cups game. These score one answer per clip; we read a released predictor's state at every step.

\paragraph{Models with built-in permanence.} ADEPT \citep{smith2019modeling} maintains probabilistic object states through occlusion by construction. PLATO \citep{piloto2022intuitive} learns violation-of-expectation responses from object-centric inputs; OPNet \citep{shamsian2020learning}, PermaTrack \citep{tokmakov2021learning} and its self-supervised successor \citep{tokmakov2022object} train trackers to survive full occlusion; and slot models \citep{locatello2020object,kipf2022savi,wu2023slotformer} and Loci-Looped \citep{traub2024loci} carry identity in dedicated slots. These build permanence in or train for it; we ask whether a general video predictor acquires it on its own, and what changes it.

\paragraph{Video world models.} V-JEPA \citep{bardes2024vjepa,assran2025vjepa2} predicts the latent features of masked tubelets from visible ones, following the JEPA programme \citep{lecun2022path}. VideoMAE \citep{tong2022videomae} reconstructs masked pixels, and Cosmos-Predict1 \citep{nvidia2025cosmos} predicts the next discrete video token. V-JEPA's masks span the whole clip by design: \citet{bardes2024vjepa} find that masks of limited temporal extent make the task easier and cost downstream accuracy (causal masks lose 3.8 points on Something-Something~v2 in their Table~4), and \citet{tong2022videomae} mask tubes against temporal leakage; Section~\ref{sec:objective} asks what that choice leaves of a hidden object. SALT \citep{li2026salt} trains a student against a frozen teacher, as our real-hand intervention does. Counterfactual inputs to a masked predictor served \citet{bear2023cwm} to extract scene structure; we use them to read a belief, with the controls that probing work asks for \citep{hewitt2019probes,li2023emergent}.

\section{Reading a Belief out of a Frozen Predictor}
\label{sec:method}

\begin{figure}[t]
\centering
\includegraphics[width=\linewidth]{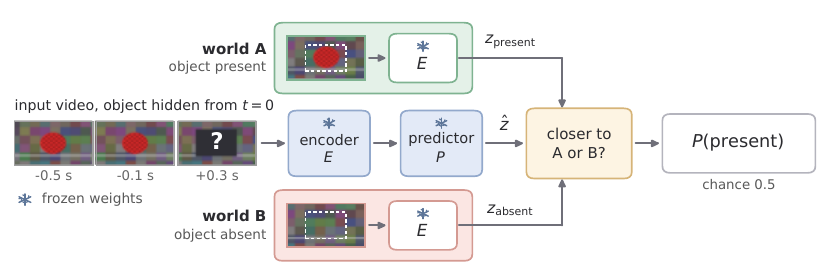}
\caption{The belief probe reads object permanence out of a frozen predictor. Once the object is hidden, its region is dropped from the encoder input and the frozen predictor $P$ infers the missing tokens; the prediction $\hat z$ is compared with the frozen encoder's representation of two worlds that are pixel-identical except inside that region. The fraction of decisions for world A is $\pres$: one value per 0.27\,s tubelet, chance 0.5.}
\label{fig:probe}
\end{figure}

If a masked video predictor maintains a belief that an occluded object is still there, that belief must be present in its prediction for the occluded region; we read it out by asking which of two possible worlds the prediction resembles more (Figure~\ref{fig:probe}).

\subsection{Minimal pairs}
We render 256$\times$256 scenes at 7.5\,fps: a textured background, $K\in\{0,2,4\}$ moving distractor rectangles, and a ball of radius $R\in\{4,8,16,32\}$\,px moving along a row at constant speed. At the event frame the ball either continues (\emph{present}) or vanishes (\emph{absent}); the worlds are pixel-identical before the event and differ only inside the ball's trajectory band after the event. A \emph{stationary} variant stops the ball four frames early, so keeping it needs no extrapolation. A checker texture removes the flat-object artefact of pixel readouts (Appendix~\ref{app:videomae}). Five seeds per cell give 60 decisions per point.

\subsection{Readout}
\label{sec:readout}
Let $E$ be the released encoder and $P$ the released predictor, both frozen. We encode a 32-frame window that starts four frames before the event, drop the tokens of the trajectory band from the event onwards, and let $P$ infer them from the remaining tokens: $\hat z = P(E(x_{\text{visible}}))$. The same $E$ encodes each full world to give references $z_{\text{present}}$ and $z_{\text{absent}}$ for the dropped tokens. V-JEPA~2 regresses its predictor onto an exponential moving average (EMA) of the encoder during pretraining; we use the released encoder for context and references, and every control is read on the same axis.

We report two quantities per tubelet slot, from the tokens of the object's expected position in that slot (a box of $R+8$\,px around it). The decision $\pres$ is the fraction of the 60 scenes whose prediction is nearer $z_{\text{present}}$ in mean per-token $L_1$, a proportion over scenes rather than a probability held by the model; chance is 0.5, but the floor for a model without memory is the matched control of Section~\ref{sec:controls}, since a prediction filled with background lands on the absent side. We call the last time point at which $\pres\ge0.5$ while its control is below 0.5 the carry-over, quoted as the start time of that tubelet. Because a regression predictor under uncertainty moves toward whichever reference is closer to its conditional estimate, a decision alone cannot say how much of the object survives, so we also report the signed distance
\begin{equation}
e = \frac{\lVert \hat z - z_{\text{absent}}\rVert_1 - \lVert \hat z - z_{\text{present}}\rVert_1}{\lVert z_{\text{present}} - z_{\text{absent}}\rVert_1},
\label{eq:evidence}
\end{equation}
a relative surprise in units of the reference separation: $e>0$ means the prediction is nearer the present world and $e=0$ that it is equidistant. Because both worlds share every visible token, the decision is the pairwise violation-of-expectation comparison of \citet{riochet2022intphys} restricted to the hidden tokens and resolved per tubelet. One tubelet spans two frames, 0.267\,s; each decision is timed by the start of its tubelet, so the first is at 0\,s and the last at 3.47\,s. The encoder never sees the hidden tokens; the only route by which the object can reach $\hat z$ is the predictor's use of the past.

Each time point pools 60 scenes, one per radius, distractor level and seed, so the 60 decisions behind a value of $\pres$ are independent (binomial 95\% half-width 0.13 at 0.5); successive time points share scenes, and comparisons with a control use a paired $t$-test on $e$ over the shared scenes.

\subsection{Controls and readout variants}
\label{sec:controls}
\emph{Never visible / never rendered.} The band is dropped in every frame, or no object is drawn under the main mask, so the model must answer absent; this bounds the readout's bias toward present (Table~\ref{tab:curves}).
\emph{Three worlds.} A \emph{static} world, the ball stopped where last seen, is added, and we report which of the three (\emph{moving}, \emph{static}, \emph{gone}) the prediction is nearest to over the hidden band.
\emph{Reveal at the end.} The band stays visible for the last four frames, showing either world, and we read the gap between occlusion and reveal.
\emph{Visible container.} An opaque box closes over the ball, moves along the trajectory or stays put, and opens after 0.53--2.93\,s; the readout is at the opening slot. Its bias control, the \emph{never-shown box}, is a box that never contained a ball. Its \emph{exit} control is a ball on the same row that leaves the frame exactly when the box closes.
\emph{Real hands.} From egocentric video with MediaPipe hand landmarks \citep{zhang2020mediapipe} we extract open$\rightarrow$closed$\rightarrow$open events, composite a coin on the open palm before the hand closes, and read the decision at re-opening in three worlds: the coin is revealed (\emph{coin-in}), it was shown and taken away 0.8\,s before the hand closed (\emph{removed}), or it was never shown (\emph{never}). The mask covers the hand from re-opening on (Appendix~\ref{app:training}).

\subsection{Probes for other paradigms}
\emph{Pixel reconstruction.} For VideoMAE \citep{tong2022videomae} we drop the same band and let the decoder reconstruct its per-patch normalised pixels; the decision compares the reconstruction with each world's patch targets.
\emph{Autoregressive.} Cosmos-Predict1-4B \citep{nvidia2025cosmos} predicts discrete tokens of a tokenizer that compresses time 8$\times$ and space 16$\times$, so one latent chunk spans 1.07\,s. Teacher forcing feeds each world its own tokens, and after the first differing chunk the present sequence has already shown the model the object, so we score that chunk alone and vary the demand by how long the ball was hidden in a visible box before it (1.07 or 2.13\,s). Both sequences share the absent world's prefix, and clips are tokenised at 256$\times$256, below the model's training resolution, so we interpret only within-scene differences of log-likelihood (Appendix~\ref{app:cosmos}). Ball tokens are less predictable than box tokens, which biases the raw log-likelihood toward absent, so we report memory as the evidence (log-likelihood of the present minus the absent tokens in the box region) minus that of the never-shown-box or exit control, per scene; Appendix~\ref{app:cosmos} gives the re-implemented forward pass and its check against natural-video log-likelihoods.

\subsection{Training interventions}
\label{sec:train}
All start from the released encoder and, except (v), the released predictor; clips are 32 frames at 7.5\,fps, 16 for (i); Table~\ref{tab:interventions} summarises them.

(i)~\emph{Temporally-partial masks on natural video}: the predictor alone is trained for 20k steps with a mixture of masks in which a spatial block is hidden from a random slot to the end of the clip, alongside causal masks (every token after a random frame) and tube masks (a spatial block in every frame), on 59k clips of egocentric and robot manipulation video (EgoDex, \citealp{hoque2025egodex}; EPIC-KITCHENS-100, \citealp{damen2022rescaling}; DROID, \citealp{khazatsky2024droid}), with an $L_1$ loss under a learned per-token Laplace scale \citep{kendall2017uncertainties} in place of the released loss.
(ii)~\emph{Synthetic container curriculum}: the predictor alone is trained with the loss of (i) for 3{,}000 steps on scenes rendered on the fly with random box and background appearance, in which a ball enters a box that later opens; the mask covers the box interior after opening.
(iii)~\emph{Mask-kind control}: the same renderer without container masks and one mask kind per run.
(iv)~\emph{Continued pretraining with an EMA target}: encoder and predictor are trained for 16k steps on the clips of (i) or on 2{,}435 internet-style clips (the FunQA humour and creative splits, \citealp{xie2024funqa}), the target encoder an EMA (0.999) of the context encoder, with tube masks only (the released mask kind) or a mixture that adds temporally-partial, causal and hand-anchored masks (centred on the detected hand) and 10\% synthetic container batches (Appendix~\ref{app:training}).
(v)~\emph{Real-hand curriculum}: starting from a predictor already trained through (i), a container curriculum and hand-anchored masks (Appendix~\ref{app:training}), the context encoder and the predictor are trained, the encoder against a frozen target, for 8{,}000 steps on batches of which 35\% are real-hand three-world clips built from 908 closing events, 20\% synthetic container scenes and the rest natural video with half of the masks hand-anchored; two EPIC-KITCHENS participants (P27, P28) are held out of training. Seeds: three for (ii) and its control curricula on ViT-H, two for the tube-mask run of (iv) on the clips of (i) and for (v), one otherwise; three recipe variants of (i) share its carry-over (Appendix~\ref{app:controls}). Appendix~\ref{app:training} lists every hyperparameter.

For IntPhys-2019 we use the fixed-context variant of \citet{garrido2025intuitive}: 7.5\,fps, two context frames, a 14-frame horizon, the prediction error averaged over windows as surprise (for predictors trained with the scale of (i), its negative log-likelihood), and pairwise accuracy over the 360 possible/impossible pairs of the dev set; their main protocol takes the minimum over context lengths; Appendix~\ref{app:review} reports both.

\section{Results}

\subsection{Released models lose a hidden moving object under a dense context}
\label{sec:curves}
\begin{figure}[t]
\centering
\includegraphics[width=\linewidth]{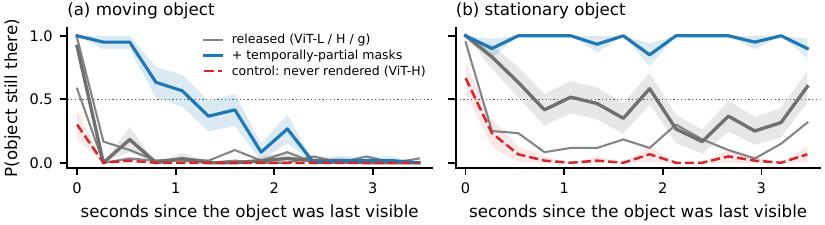}
\caption{Scale does not buy permanence: on a moving object under the main query all three released models cross to \emph{absent} after one tubelet, and a predictor trained with temporally-partial masks keeps a moving object for 1.1\,s and a stationary one to the end of the window. Decision that the hidden object is still there, by seconds since it was last visible. Grey: the three released scales (thick: ViT-H); red dashed: the never-rendered control for ViT-H (the floor for no memory); dotted: chance. Shaded: binomial 95\% intervals over 60 scenes.}
\label{fig:curves}
\end{figure}

With the rest of the frame visible (a dense context), the released V-JEPA~2 predictor keeps a hidden moving object for one tubelet and then leans absent. Figure~\ref{fig:curves}a and Table~\ref{tab:curves} (Appendix~\ref{app:tables}) give the curve for ViT-H: $\pres=0.92$ at 0\,s and 0.00 at 0.27\,s, against a geometry-matched control, the object never rendered, of 0.30 and 0.00. ViT-L (0.58, 0.00) and ViT-g (0.98, 0.17, 0.10) behave the same way, so scale does not help under this query. A stationary object is kept in part: from 0.53\,s on, ViT-H reads it as present in 41\% of scenes on average (17--63\% at single time points), ViT-L in 16\% (3--32\%) and ViT-g in 13\% (3--27\%), against controls of at most 7\% (Table~\ref{tab:curves}; ViT-L and ViT-g controls in Appendix~\ref{app:controls}), while a copy of the last visible tokens reads 97--98\% on average; the released predictor keeps less of a stationary object than the frame it last saw. The three-way readout (Figure~\ref{fig:threeway}, Appendix~\ref{app:controls}) puts the prediction nearest the \emph{gone} world in 60--97\% of decisions from 0.8\,s on. Frame rate changes little (Appendix~\ref{app:review}); context density does. With the block ahead of the ball masked in every frame (V-JEPA's own tube geometry) and a 2.1\,s visible approach, ViT-L, ViT-H and ViT-g read 0.08, 0.52 and 0.33 one tubelet after entry and at most 0.30 afterwards; hiding a random 90\% of the remaining context as well, pretraining's masking ratio, leaves ViT-L and ViT-g there (0.12, 0.18) but lets ViT-H keep the ball to 1.1\,s (1.00, 0.87, 0.95, 0.87, 0.90 against a control of 0.70 then 0.02). Predictors trained on synthetic scenes (Section~\ref{sec:install}) keep it through the tube for the whole window (Appendix~\ref{app:review}).

The signed distance of Equation~\ref{eq:evidence} shows what the decision hides. For ViT-H it moves from $+0.014$ at 0\,s to $-0.030$ at 0.27\,s, so a swing of 0.04 in $e$ flips $\pres$ from 0.92 to 0.00. The prediction does not land on the never-rendered control, which reads $-0.094$ there, and the gap holds at every step to 3.47\,s (paired $t$-test over the 60 scenes, $p<10^{-18}$ at each step): a trace survives, too weak to carry the decision. Because $e$ is compressed by prediction error in every direction (at 0\,s a copy of the last visible tokens decides \emph{present} in 98\% of scenes yet reads $+0.105$), we also project the prediction onto the axis between the worlds: at 0.53\,s it lies 0.29 of the axis above the control where the copy lies 0.73, and from 0.27\,s on it stays below the midpoint, decreasing slowly (0.43 to 0.33 by 2.93\,s; Appendix~\ref{app:review}). ``Lost'' describes the decision, not the information. This model scores 84.2\% on IntPhys-2019; Section~\ref{sec:install} returns to what that score measures.

Training the predictor alone with temporally-partial masks on natural video (intervention (i), v2 in the tables) extends the moving-object decision to 1.1\,s and the stationary one to the end of the window (Table~\ref{tab:curves}, rows v2). The three-way readout explains the limit: from 0.8 to 2.4\,s the trained predictor's decisions go mostly to the \emph{static} world (0.77 at 0.8\,s, \emph{moving} 0.18), and \emph{gone} is the most frequent answer only at 2.7--3.2\,s. Natural video with these masks teaches that an occluded thing is still where it was, not that it kept moving.

\subsection{The predictor inpaints locally under a dense context}
\label{sec:reveal}
The released predictor does not interpolate across a gap even when both ends are visible. With the last four frames of the band left visible and showing the ball, the belief in the gap stays where it is without a reveal (0.00--0.18 from 0.27 to 2.67\,s), and the only slot the object's reveal moves is the last hidden one, adjacent to it, from 0.00 to 0.15 (Figure~\ref{fig:reveal}); a reveal of the empty band lowers the first tubelet from 0.92 to 0.72 and leaves the gap at most 0.07. Tube masking at a low masking ratio hides a block in every frame while its neighbours stay visible, as the band's neighbours do here, so a prediction can be built from the same instant rather than from the past, a reading the sparse-context query of Section~\ref{sec:curves} supports but does not isolate as a cause (Appendix~\ref{app:scope}); Section~\ref{sec:objective} shows that changing the masks and the data changes the reach.

Training changes the reach in both directions. The predictor trained with temporally-partial masks reads 0.32--1.00 across the gap when the reveal shows the ball and falls to 0.05 by 0.8\,s when it shows the ball absent (0.63, 0.57, 0.37 with no reveal): it carries the object forward and revises from the future. The synthetic curricula go to the other extreme: with the reveal showing the ball absent, the container-curriculum predictor still reads 0.65--1.00 until 1.6\,s and the real-hand model 0.80--1.00 throughout (never-visible controls 0.32 at 0\,s, then at most 0.02).

\begin{figure}[t]
\centering
\includegraphics[width=\linewidth]{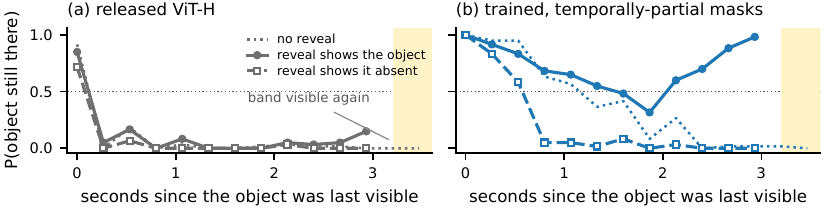}
\caption{The released predictor fills in only from its immediate neighbours: revealing the object in the last four frames moves its prediction only in the adjacent tubelet, while the trained predictor carries the object forward and revises it from the future. Decision in the gap when the reveal shows the object (solid) or its absence (dashed), against no reveal (dotted).}
\label{fig:reveal}
\end{figure}

\subsection{The information is in the encoder and the predictor barely uses it}
\label{sec:encoder}
The deficit is not one of perception. A logistic-regression probe on the frozen encoder's tokens decodes the ball's presence at 1.00, down to 4\,px. For a visible box that has closed over the ball and moved on, the box's own tokens keep the contents linearly decodable at 0.91--1.00 for the full 3.5\,s after the event (the box never opens in these clips). This is decodability within the one window of bidirectional attention the predictor also sees, not a state maintained across windows.

The predictor, given matched scenes in which the box opens after 0.53--2.93\,s, reads 0.05 at 0.53\,s, 0.02 at 1.07\,s and 0.00 afterwards for a moving box, and 0.05, 0.03, 0.17 and 0.08 for a stationary one (Table~\ref{tab:container}). Its signed distance $e$ sits above the never-shown-box control's by $+0.008$ at 0.53\,s and $+0.003$ at 2.93\,s (paired $p<10^{-14}$), against $+0.13$ for the container curriculum of Section~\ref{sec:install} at a decision of 1.00. The encoder's own probe, applied to the predictor's output, reads the ball inside the box in 2\% of scenes at 0.53\,s and in none afterwards (Appendix~\ref{app:review}), so the gap is not made by the readout. A probe fitted on the predictor's output itself separates the main condition from either control at 0.99--1.00 at every delay (0.93--1.00 against the exit control at the other scales): the context reaches the output, and what is missing is the presence feature at the object's position.

A real closing hand moves the limit to the other side: there the encoder itself loses the coin (decodable from the hand's tokens at 0.75--0.89 while the palm is open, 0.56--0.59 once closed), which is why the real-hand intervention trains the encoder as well (Section~\ref{sec:install}).

\subsection{The deficit is not intrinsic to latent prediction}
\label{sec:objective}
Neither the mask kind nor the video source withholds continuation. On synthetic rigid-motion scenes without container masks, a predictor acquires the same 1.3--1.6\,s of carry-over under each single mask kind we tried, keeps a stationary object for the whole window and keeps no ball inside a visible box (stationary 0.77--1.00 at every step, ball in a box at most 0.18; Table~\ref{tab:maskkind}). All three raise IntPhys-2019 above the 84.2\% baseline (87.8--91.9\%).

Natural video separates two habits. Continuing with the released mask kind, tube masks, for 16k steps, with an EMA target and the encoder trainable, produces 1.3 and 1.1\,s of moving-object carry-over on the manipulation video of (i) in two seeds and 1.6\,s on internet-style video (Figure~\ref{fig:objdata}, Table~\ref{tab:ema}), where the released ViT-L has none beyond the first tubelet (never-visible controls 0.12 and 0.08 at 0\,s, then 0.00; Appendix~\ref{app:controls}). A stationary hidden object is not kept by these runs: 0.28, 0.37 and 0.15 at 1.07\,s, close to the released ViT-L's 0.12. Which recipe difference carries the effect, and at what cost, is unmeasured (Appendix~\ref{app:scope}). The mask mixture changes the form. With temporally-partial masks in it the model freezes the object where it was last seen (0.82 \emph{static} at 0.8\,s), holds a stationary object at 1.00 to the end of the window and, from the 10\% container batches, keeps a ball inside a box at 1.00 against both controls (Appendix~\ref{app:controls}), at the price of a shorter moving-object curve (0.50 at 0.53\,s). With two context frames IntPhys-2019 moves from 61.4\% to 62.5--64.2\% with tube masks and to 53.1\% with the mixture; under the minimum rule all four rise from 45.0\% (to 59.2--63.1\%; Table~\ref{tab:protocol}). Tube masks on natural video thus produced extrapolation in two of three runs (three-world column of Table~\ref{tab:ema}), and the mixture with temporally-partial masks produced persistence in place.

\begin{figure}[t]
\centering
\includegraphics[width=\linewidth]{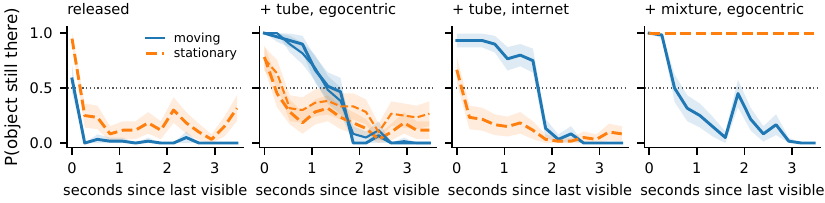}
\caption{Continued training with tube masks produces moving-object carry-over on either video source but does not keep a stationary object; with the mask mixture the object freezes instead. Belief for a hidden moving (solid) and stationary (dashed) object, ViT-L, one panel per training condition; the two lines in the second panel are two seeds. Shaded: binomial 95\% intervals.}
\label{fig:objdata}
\end{figure}

\subsection{Pixel-reconstruction and autoregressive models under matched probes}
\label{sec:paradigms}
\begin{figure}[t]
\centering
\includegraphics[width=\linewidth]{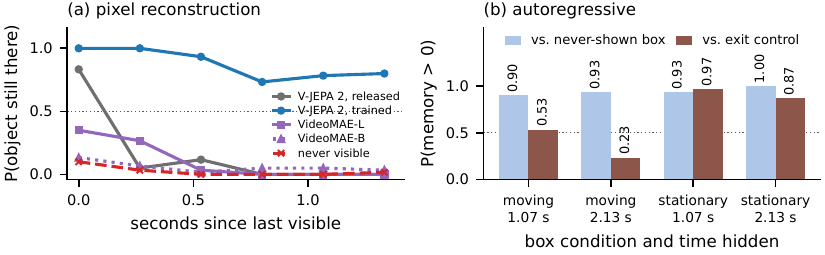}
\caption{VideoMAE keeps almost nothing; Cosmos keeps a stationary hidden object and not a carried one. (a) Belief curves for the textured ball under the pixel readout (VideoMAE) next to the latent readout (V-JEPA~2, released and after (i)). (b) Cosmos-Predict1-4B: fraction of 30 scenes expecting the ball at the reveal, against two controls; only a stationary box survives the exit control.}
\label{fig:paradigms}
\end{figure}

VideoMAE keeps almost nothing and Cosmos keeps only what stays put. VideoMAE-L reconstructs a hidden 32\,px ball for 0.3\,s and the base models for none (Appendix~\ref{app:videomae}). Cosmos-Predict1-4B, which only ever sees the past, shows the static-versus-carried split inside one model (Figure~\ref{fig:paradigms}b, Table~\ref{tab:cosmos}). A ball in a box that stays put is expected at a reveal 1.07 and 2.13\,s later: memory against the exit control is $+20.7$ and $+18.7$ nats, positive in 97\% and 87\% of scenes. A ball carried inside a moving box is not: $-2.7$ and $-10.3$ nats (95\% intervals [$-9.0$, 3.0] and [$-17.4$, $-3.6$]), positive in 53\% and 23\%. Against the never-shown box alone the moving box also looks remembered ($+20.1$ nats); the exit control shows that to be the expectation of a ball somewhere in the scene, not in the box. Tracking an object inside a moving container is a habit neither the released V-JEPA~2 nor Cosmos acquires from natural video, and VideoMAE keeps almost nothing even without a container.

\subsection{What training installs}
\label{sec:install}
\begin{figure}[t]
\centering
\includegraphics[width=\linewidth]{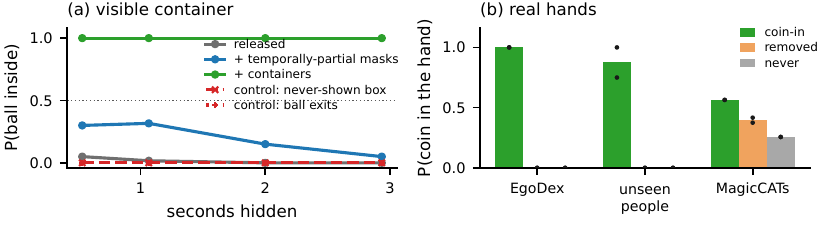}
\caption{Training installs what the released model lacks. (a) Three thousand predictor-only steps take a ball inside a moving box from at most 0.05 to 1.00 at every hidden duration, with the never-shown-box and exit controls at 0.00. (b) Training the context encoder on real hands, intervention (v), separates coin-in, removed and never on held-out clips and on two participants never seen in training (bars: mean of two seeds; dots: each seed).}
\label{fig:install}
\end{figure}

\paragraph{Containers.} Three thousand predictor-only steps on synthetic container scenes take the belief that a ball is inside a visible box from at most 0.17 to 1.00 at every hidden duration up to 2.93\,s, for a moving and a stationary box, with both controls at 0.00--0.02 (Figure~\ref{fig:install}a, Table~\ref{tab:container}). The curriculum trains on the probe's own renderer, so 1.00 there is expected; over natural-video backgrounds, the one held-out test, it reads 0.85--0.97; what is installed is a prior that a reveal of the empty band does not revise (Section~\ref{sec:reveal}), at an unmeasured cost elsewhere (Appendix~\ref{app:scope}). The same curriculum raises IntPhys-2019 from 84.2\% to 93.3\% for ViT-H, from 61.4\% to 64.4\% for ViT-L and from 54.2\% to 75.0\% for ViT-g (the released ViT-L and ViT-g score far below ViT-H under this protocol, which we do not explain), and for ViT-H the gain survives seeds and readouts (Table~\ref{tab:intphys}); training only the loss's per-token scale head with the predictor frozen gives 82.8\%, so the gain is in the predictor, not the scoring.

\paragraph{What the benchmark measures.} The gain is not containment: without the container masks the curriculum (\emph{moving} below) gives 93.9--94.7\% over three seeds. How it divides between motion, persistence and the rendered domain depends on the scoring rule. With two context frames, over three training seeds each, a curriculum in which nothing moves (\emph{static}) gives 93.8\% and one in which every object is redrawn at a random position in every frame (\emph{non-persistent}) 88.7\%, 5.1 points lower (95\% interval [1.9, 8.4]; Appendix~\ref{app:review}). With the minimum surprise over context lengths, the rule of \citet{garrido2025intuitive}, our implementation of which does not reproduce their published numbers and serves only to compare our own checkpoints (Appendix~\ref{app:review}), the moving curriculum leads (93.3\%) and the static and non-persistent ones tie (90.6 and 91.0\%), whereas the probe ties all three on a hidden stationary ball and ranks the static one last on a moving ball. The natural-video mixture of Section~\ref{sec:objective} likewise changes side with the rule, and the real-hand recipe below lowers the score under both. A gain on the benchmark is therefore not by itself evidence of permanence, and for these checkpoints its accuracy does not track the belief the probe reads.

\begin{table}[t]
\centering\footnotesize
\caption{IntPhys-2019 dev set, pairwise accuracy (\%) under the protocol of Section~\ref{sec:train}. Two context frames; $L_1$ readout: surprise from the released $L_1$ loss instead of the trained scale head; seeds and other scoring rules in Table~\ref{tab:protocol} ($^\dagger$RTX~5090 machines, Appendix~\ref{app:training}; --: not run).}
\label{tab:intphys}
\begin{tabular}{lccc}
\toprule
 & ViT-L & ViT-H & ViT-g \\
\midrule
released (baseline) & 61.4 & 84.2 & 54.2 \\
+ container curriculum, 3{,}000 predictor-only steps & 64.4 & 93.3 & 75.0 \\
\quad two further seeds / $L_1$ readout & -- / 64.4 & 93.1, 93.3 / 92.8 & -- / 73.9 \\
\quad no container masks (3 seeds for ViT-H)$^\dagger$ & 67.8 & 94.3 & 77.5 \\
\quad\quad static / non-persistent$^\dagger$ & -- / 66.7 & 93.8 / 88.7 & -- / 70.0 \\
\bottomrule
\end{tabular}
\end{table}

\paragraph{Real hands.} The released encoder drops the coin once a real hand closes (Section~\ref{sec:encoder}), and a predictor trained alone on two-world real-hand clips learns a shortcut: coin-in reads 0.75 and removed reads 1.00, so a hand that held a coin reveals a coin. Training the context encoder with three worlds and container batches removes the shortcut (Figure~\ref{fig:install}b, Table~\ref{tab:hands}). It separates coin-in, removed and never at 1.00 / 0.00 / 0.00 on 8 held-out EgoDex events and at 0.75--1.00 / 0.00 / 0.00 on 4 events of two EPIC-KITCHENS participants excluded from training (3 and 2 with a removed world), a sample that fixes the direction and not the magnitude. The closed hand's own tokens now carry the coin (after closing 0.74--0.85, against 0.59 before training); on the MagicCATs hands \citep{ozono2021magiccats} the transfer is partial (0.56 / 0.38--0.42 / 0.26 over 39 events). The price is IntPhys-2019: 65.8\% for the encoder-trained model with containers and 27.2\% and 30.6\%, below chance, for the two seeds of (v).

\section{Conclusion and Limitations}
The released V-JEPA~2 encoders represent an object they can see; the predictors keep little of it once it is hidden. With the rest of the frame visible they cross to \emph{absent} for a moving object within a tubelet or two (only ViT-H, at pretraining's 90\% masking ratio, keeps it to 1.1\,s), keep a stationary one in a minority of scenes, and leave in the encoder information they barely touch: where a linear probe reads the object at ceiling, the same probe finds it in the predictor's output in 2\% of scenes once a box has closed over it. What persists changes with the masks and the data rather than with latent prediction itself; V-JEPA~2 and Cosmos fail alike on an object carried inside a moving container, three thousand predictor-only steps close that gap for V-JEPA~2, and the IntPhys-2019 gain they bring does not follow the belief the probe reads. The limits are those of the evidence: one model family's pretraining predictor (not the action-conditioned or 2.1 models), one renderer and no two-world test on real occlusion video, curricula trained on the probe's own distribution, thresholded decisions over 60 scenes per point, and an IntPhys-2019 implementation that does not reproduce published minimum-rule numbers; Appendix~\ref{app:scope} states what each does and does not show and what would change it.

\subsection*{AI use statement}
In this work, we used generative AI tools, a large language model assistant, to help develop the conceptual framing, propose and refine hypotheses, design experiments and controls, implement the methods, clean and reformat the datasets (video decoding, hand-event extraction and data manifests), interpret results, and translate the authors' notes into English. We have not used generative AI tools to generate synthetic data sets: the synthetic scenes are drawn by procedural rendering code that the assistant helped write, and no generative model produced any image, video or data point. Formulating mathematical claims, providing ingredients for proofs, writing proofs, and qualitative or thematic data analysis are not applicable to this work. Additionally, we used generative AI tools to write and edit software code, suggest experimental parameters, create the figures, search for, identify and summarise related work, suggest the structure, title and keywords of the paper, draft its text, edit it for readability and format the references. We have reviewed all AI-assisted work: the authors decided which experiments to run and which claims to make; every reference was checked against its DOI, arXiv or publisher record; the reported numbers were checked against the logged outputs of the runs; and defects in model-written code were caught and corrected before the numbers reported here, among them a data manifest with relative paths that trained one run on blank clips, a cross-validation split that changed with the library version, a first version of the autoregressive probe that scored chunks after the object had been shown, and a table script that reported the best of several IntPhys settings instead of the stated one. We take responsibility for the final content of this work, including text, claims or artifacts produced with the aid of generative AI.

\subsection*{Ethics statement}
All natural video used here comes from existing published datasets and we collected none. EgoDex, EPIC-KITCHENS-100, DROID, MagicCATs, MMC and FunQA were obtained from their authors or distributors under their own terms and used for research; MagicCATs and MMC are published stimulus sets distributed by their authors for research, and both carry human ratings that we use only in aggregate. Frames from these datasets show identifiable people. We do not redistribute them, we report no participant-level result beyond held-out group accuracy, and the coins composited onto real hands are synthetic overlays that change no person's appearance or action. Holding out participants P27 and P28 is a generalisation test, not a judgement about those recordings.

\subsection*{Reproducibility statement}
The probe, renderer, training and evaluation scripts, together with the manifests and job scripts that produced every number in this paper, will be released. Section~\ref{sec:method} defines the readout and every control; Appendix~\ref{app:training} lists all hyperparameters, the values tried for every hyperparameter we varied, library versions and the compute used; Appendix~\ref{app:data} describes preprocessing and the held-out splits; Appendix~\ref{app:review} states the IntPhys-2019 protocol, its relation to that of \citet{garrido2025intuitive}, and the scene-level bootstrap. The human-rated magic datasets are available from their authors under their own terms.

\subsection*{Acknowledgements}
Experiments presented in this work were carried out using the CIT-TUM-HN cluster at TUM Campus Heilbronn.

\bibliographystyle{iclr2027_conference}
\bibliography{references}

\appendix
\setlength{\abovetopsep}{3pt}

\section{Full Result Tables}
\label{app:tables}
\begin{table}[h]
\centering\small
\caption{The five training interventions of Section~\ref{sec:train}, all starting from the released encoder and, except (v), the released predictor; the IntPhys-2019 gain is for ViT-H with two context frames; (v) starts from a predictor already trained through (i), a container curriculum and hand-anchored masks (Appendix~\ref{app:training}).}
\label{tab:interventions}
\setlength{\tabcolsep}{3pt}
\begin{tabular}{llll}
\toprule
Intervention & Data & Trained & Installs \\
\midrule
(i) temporally-partial masks & natural video & predictor, 20k & object kept where last seen \\
(ii) container masks & synthetic & predictor, 3k & containment prior; IntPhys $+9.1$ \\
(iii) single mask kind & synthetic & predictor, 3k & carry-over only \\
(iv) tube masks, EMA & video of (i) / internet & enc. + pred., 16k & 1.1--1.6\,s carry-over \\
(v) three-world coin & real hands & ctx.\ enc. + pred., 8k & coin belief, unseen people \\
\bottomrule
\end{tabular}
\end{table}
\begin{table}[h]
\centering\small
\caption{Belief that a hidden object is still there, $\pres$, by seconds since it was last visible (60 decisions per cell; chance 0.5; the matched control is the floor for no memory). The released models drop to at most 0.17 at 0.27\,s and at most 0.18 afterwards on a moving object; the predictor trained with temporally-partial masks (v2) keeps a moving object for 1.1\,s and a stationary one for the whole window (bold: its values at or above 0.5). --: not evaluated.}
\label{tab:curves}
\begin{tabular}{llccccccc}
\toprule
Object & Model & 0 & 0.27 & 0.53 & 0.80 & 1.07 & 1.33 & 1.60 \\
\midrule
moving & ViT-L released & 0.58 & 0.00 & 0.03 & 0.02 & 0.02 & 0.00 & 0.02 \\
moving & ViT-H released & 0.92 & 0.00 & 0.18 & 0.00 & 0.03 & 0.00 & 0.00 \\
moving & ViT-g released & 0.98 & 0.17 & 0.10 & 0.02 & 0.03 & 0.02 & 0.10 \\
moving & ViT-H, never rendered (control) & 0.30 & 0.00 & 0.02 & 0.00 & 0.00 & 0.00 & 0.00 \\
moving & ViT-H + temporally-partial (v2) & \best{1.00} & \best{0.95} & \best{0.95} & \best{0.63} & \best{0.57} & 0.37 & 0.42 \\
moving & v2, never visible (control) & 0.18 & 0.00 & 0.00 & 0.00 & 0.00 & 0.00 & -- \\
\midrule
stationary & ViT-L released & 0.95 & 0.25 & 0.23 & 0.08 & 0.12 & 0.12 & 0.18 \\
stationary & ViT-H released & 1.00 & 0.83 & 0.63 & 0.42 & 0.52 & 0.47 & 0.35 \\
stationary & ViT-g released & 1.00 & 0.12 & 0.27 & 0.10 & 0.03 & 0.12 & 0.25 \\
stationary & ViT-H, never rendered (control) & 0.67 & 0.23 & 0.07 & 0.02 & 0.00 & 0.02 & 0.00 \\
stationary & ViT-H + temporally-partial (v2) & \best{1.00} & \best{0.90} & \best{1.00} & \best{1.00} & \best{1.00} & \best{0.93} & \best{1.00} \\
\bottomrule
\end{tabular}
\end{table}

\begin{table}[h]
\centering\small
\caption{Mask kind does not decide carry-over on synthetic rigid-motion scenes; belief by seconds since last visible. Predictors trained on the same rendered scenes with a single mask kind each (3{,}000 steps, no container masks) acquire the same moving-object belief, keep a stationary object (0.77--1.00 at every step; last step 0.82 / 0.95 / 0.92 for causal / partial / tube) and none acquire containment (ball in a visible box at 0.53\,s hidden: 0.05 / 0.18 / 0.18, same order); IntPhys-2019 (\%, two context frames) rises under all three; in these scenes an object that enters a hidden block keeps moving, so predicting the block rewards continuation under each of these masks. These runs have no never-visible control of their own (Appendix~\ref{app:controls}).}
\label{tab:maskkind}
\begin{tabular}{lcccccccc}
\toprule
Training masks & 0 & 0.27 & 0.53 & 0.80 & 1.07 & 1.33 & 1.60 & IntPhys-2019 \\
\midrule
released ViT-H (none) & 0.92 & 0.00 & 0.18 & 0.00 & 0.03 & 0.00 & 0.00 & 84.2 \\
tube only (released mask kind) & 1.00 & 0.98 & 0.97 & 0.77 & 0.77 & 0.58 & 0.53 & 91.1 \\
temporally-partial only & 1.00 & 0.97 & 0.97 & 0.68 & 0.75 & 0.67 & 0.50 & 87.8 \\
causal only & 1.00 & 1.00 & 0.97 & 0.72 & 0.72 & 0.58 & 0.42 & 91.9 \\
\bottomrule
\end{tabular}
\end{table}

\begin{table}[h]
\centering\small
\caption{Continued training with tube masks for 16k steps produces carry-over on the video of (i) (two seeds) and on internet-style video; with the mask mixture the model freezes the object instead. ViT-L with an EMA target; IntPhys-2019 (\%) with two context frames; bold: $\pres\ge0.5$ after training, and the best IntPhys-2019 score. Never-visible controls in Appendix~\ref{app:controls}.}
\label{tab:ema}
\setlength{\tabcolsep}{3pt}
\begin{tabular}{lcccccc}
\toprule
ViT-L, 16k & \multicolumn{3}{c}{moving, $\pres$ at} & stat. & 3 worlds at 0.80\,s & IntPhys- \\
\cmidrule(lr){2-4}
 & 0.27\,s & 0.80\,s & 1.33\,s & 1.07\,s & moving / static / gone & 2019 \\
\midrule
released & 0.00 & 0.02 & 0.00 & 0.12 & 0.00 / 0.03 / 0.97 & 61.4 \\
+ tube masks, seed 1 & \best{0.97} & \best{0.90} & \best{0.52} & 0.28 & 0.45 / 0.37 / 0.18 & 62.8 \\
+ tube masks, seed 2 & \best{1.00} & \best{0.82} & 0.48 & 0.37 & 0.27 / 0.57 / 0.17 & \best{64.2} \\
+ tube masks, internet & \best{0.93} & \best{0.90} & \best{0.80} & 0.15 & 0.72 / 0.17 / 0.12 & 62.5 \\
+ mask mixture & \best{0.98} & 0.32 & 0.15 & \best{1.00} & 0.10 / 0.82 / 0.08 & 53.1 \\
\bottomrule
\end{tabular}
\end{table}

\begin{table}[h]
\centering\small\setlength{\tabcolsep}{4pt}
\caption{A ball inside a visible box: belief at the moment the box opens, by seconds hidden. The released predictor loses it within 0.5\,s (controls: never-shown box / exit); 3{,}000 predictor-only steps on synthetic containers install it at 1.00 with both controls at 0.00 (stationary-box controls, not shown: 0.00--0.02).}
\label{tab:container}
\begin{tabular}{llcccc}
\toprule
Model & Condition & 0.53 & 1.07 & 2.00 & 2.93 \\
\midrule
ViT-H released & moving box & 0.05 & 0.02 & 0.00 & 0.00 \\
ViT-H released & never-shown box (control) & 0.00 & 0.00 & 0.00 & 0.00 \\
ViT-H released & stationary box & 0.05 & 0.03 & 0.17 & 0.08 \\
ViT-H released & stationary box, controls & 0.02 / 0.03 & 0.00 / 0.00 & 0.03 / 0.07 & 0.00 / 0.00 \\
v2 & moving box & 0.30 & 0.32 & 0.15 & 0.05 \\
v2 & stationary box & 0.23 & 0.38 & 0.53 & 0.47 \\
+ container curriculum & moving box & \best{1.00} & \best{1.00} & \best{1.00} & \best{1.00} \\
+ container curriculum & stationary box & \best{1.00} & \best{1.00} & \best{1.00} & \best{1.00} \\
+ container curriculum & never-shown box (control) & 0.00 & 0.00 & 0.00 & 0.00 \\
+ container curriculum & exit (control) & 0.00 & 0.00 & 0.00 & 0.00 \\
+ container curriculum & natural background & 0.97 & 0.93 & 0.90 & 0.85 \\
\bottomrule
\end{tabular}
\end{table}

\begin{table}[h]
\centering\small
\caption{A three-world belief about closed real hands transfers to participants never seen in training. Belief that a coin is in the re-opened hand, hands closed for at most 0.6\,s; the coin-in world should read high, the other two 0.00. Clean recipe: intervention (v) as specified in Section~\ref{sec:train}, two seeds (s1 / s2); the rows above it are its development stages. Events: 8 (EgoDex), 4 (EPIC-KITCHENS) and 39 (MagicCATs), of which 3, 2 and 24 have a removed world, since removal needs the coin in view 0.8\,s before closing; bold: all three worlds separated. Event counts in Appendix~\ref{app:hands}.}
\label{tab:hands}
\setlength{\tabcolsep}{4pt}
\begin{tabular}{llccc}
\toprule
Model & Held-out set & coin-in & removed & never \\
\midrule
released ViT-H & MagicCATs hands & 0.36 & 0.42 & 0.33 \\
predictor only, 2 worlds & EgoDex & 0.75 & 1.00 & 0.00 \\
encoder trained, 2 worlds & EgoDex & 1.00 & 1.00 & 0.00 \\
encoder trained, 3 worlds & EgoDex & 1.00 & 0.33 & 0.00 \\
encoder trained, 3 worlds + containers & EgoDex & \best{1.00} & \best{0.00} & \best{0.00} \\
\midrule
clean recipe, s1 / s2 & EgoDex & \best{1.00 / 1.00} & \best{0.00 / 0.00} & \best{0.00 / 0.00} \\
clean recipe, s1 / s2 & EPIC-KITCHENS P27/P28 & \best{0.75 / 1.00} & \best{0.00 / 0.00} & \best{0.00 / 0.00} \\
clean recipe, s1 / s2 & MagicCATs hands & 0.56 / 0.56 & 0.38 / 0.42 & 0.26 / 0.26 \\
\bottomrule
\end{tabular}
\end{table}

\begin{table}[h]
\centering\small
\caption{An autoregressive model keeps a stationary hidden object and not a carried one. Cosmos-Predict1-4B, teacher-forced log-likelihood of the reveal chunk; memory is the per-scene difference between the main condition and a control (30 scenes per row); visible continuation: the ball stays in view and the chunk after the event is scored; --: not applicable.}
\label{tab:cosmos}
\setlength{\tabcolsep}{4pt}
\begin{tabular}{lccccc}
\toprule
 &  & \multicolumn{2}{c}{vs. never-shown box} & \multicolumn{2}{c}{vs. exit control} \\
\cmidrule(lr){3-4}\cmidrule(lr){5-6}
Condition & Hidden & nats & $P(>0)$ & nats & $P(>0)$ \\
\midrule
box moving & 1.07\,s & $+20.1$ & 0.90 & $-2.7$ & 0.53 \\
box moving & 2.13\,s & $+18.2$ & 0.93 & $-10.3$ & 0.23 \\
box stationary & 1.07\,s & $+35.8$ & 0.93 & \best{$+20.7$} & \best{0.97} \\
box stationary & 2.13\,s & $+38.7$ & 1.00 & \best{$+18.7$} & \best{0.87} \\
visible continuation & -- & $+153.4$ & 1.00 & -- & -- \\
\bottomrule
\end{tabular}
\end{table}

\section{Datasets and Preprocessing}
\label{app:data}
Natural videos are decoded at 7.5\,fps (IntPhys~2 at 6\,fps, Qwen3-VL labelling at 2\,fps) and resized to 256$\times$256 (squash) for V-JEPA~2 and 224$\times$224 for VideoMAE. Natural-video training uses 59k clips: EgoDex (28k clips), EPIC-KITCHENS-100 (579 videos cut into clips) and DROID (30k clips). Hand landmarks come from MediaPipe Hands with 21 points per hand; closing events require openness above 1.6 palm lengths before and after, below 1.35 during, for at least three frames. The 908 training events exclude EPIC-KITCHENS participants P27 and P28, whose 11 events form the unseen-participant set; the EgoDex held-out set contains 19 events from 1{,}500 clips withheld from every training manifest. MagicCATs contributes 166 clips with 81 detectable closing events; MMC (magic, memory and curiosity) contributes 36 videos with 99 human-rated surprise moments and 21 detectable closing events, 19 of which (from 11 videos) the probe reads. IntPhys~2 is CC BY-NC 4.0 and limited to evaluation; EPIC-KITCHENS-100 is CC BY-NC 4.0; EgoDex is CC BY-NC-ND; FunQA is CC BY-NC-SA 4.0; IntPhys-2019, DROID, MagicCATs and MMC were obtained under the terms stated by their distributors, the last two on request.
\paragraph{Renderer.} Synthetic scenes are 256$\times$256 frames on a background of 16\,px tiles with random intensities in 70--130. The ball, radius $R\in\{4,8,16,32\}$\,px, moves along the row $y=190$ from $x=28$ to $x=228$ over the clip at constant speed (3.2\,px per frame in 64-frame clips, 24\,px/s at 7.5\,fps) above a thin guide line; $K\in\{0,2,4\}$ distractor rectangles of 36--56 by 48--72\,px follow smooth sinusoidal paths in the upper region ($y\le130$) and never reach the ball's row; the visible box is an opaque 80$\times$88\,px rectangle; the checker texture alternates two reds in 4\,px cells. The event frame is the clip's midpoint and the 32-frame window starts four frames before it (64-frame windows starting 8 and 16 frames before it at 15 and 30\,fps).

\section{Training Details}
\label{app:training}
\paragraph{Predictor-only training on natural video (v2).} Released ViT-H encoder frozen; released predictor initialised from the checkpoint; 20k steps, batch 8, 16-frame windows at frame step 2, learning rate $5\times10^{-5}$ with 200 warm-up steps; masks sampled per batch from causal (30\%), temporally-partial (50\%) and tube (20\%) kinds; loss $L_1$ on encoder features with a learned per-token Laplace scale. Variants: 32-frame windows throughout, a re-weighted loss over later slots, a 24-layer predictor.
\paragraph{Synthetic container curriculum.} 3{,}000 steps, batch 12, 32 frames, loss as for v2, scenes rendered on the fly: random box colour and size, random background tile, $K$ distractors; 70\% of batches use the container mask (box interior after opening), 30\% the generic mixture; scenarios enter, miss, exit and none. Ablations: two further seeds; container-mask fraction 0 (the no-container-mask curriculum) and 1 (89.2\% on IntPhys-2019); scale head only, predictor frozen (82.8\%).
\paragraph{Mask-kind control.} As above with the container-mask fraction at 0 and a single mask kind (tube, temporally-partial or causal).
\paragraph{Continued pretraining with an EMA target.} ViT-L; context encoder trainable at learning rate $3\times10^{-5}$, predictor at $10^{-4}$, 1{,}000 warm-up steps; target encoder is an EMA of the context encoder with momentum 0.999, kept in fp32; 16k steps, batch 6 per GPU on four GPUs, 32 frames; tube-only masks, or the mixture tube 40\% / temporally-partial 40\% / causal 10\% / hand-anchored 10\% with 10\% synthetic container batches.
\paragraph{Real-hand curriculum.} ViT-H; context encoder trainable at $10^{-4}$ with the released encoder as fixed target, predictor at $5\times10^{-5}$; 8{,}000 steps, batch 8; 35\% real-hand batches (worlds coin-in 50\%, removed 25\%, never 25\%; zoom 1.6--2.2$\times$, random coin colour and size, horizontal flips; the mask covers the hand from re-opening on with a shared token budget per batch), 20\% synthetic container batches, 45\% natural video with the mixture causal 15\% / temporally-partial 25\% / tube 10\% / hand-anchored 50\%; the predictor is initialised from one trained alone for 6{,}000 steps on EgoDex and EPIC-KITCHENS clips with hand-anchored masks added to the mixture of v2 and 20\% synthetic container batches, itself initialised from a container curriculum run on top of v2; seeds 1 and 2.
\paragraph{Checkpoints.} V-JEPA~2: \path{facebook/vjepa2-vitl-fpc64-256}, \texttt{-vith-} and \texttt{-vitg-}. VideoMAE: the MCG-NJU pretraining checkpoints with decoder (base and large, Kinetics-400; base, Something-Something~v2). Cosmos: the model \path{nvidia/Cosmos-Predict1-4B} with the tokenizer \path{nvidia/Cosmos-Tokenize1-DV8x16x16-720p}. Labelling of the magic clips: \path{Qwen/Qwen3-VL-8B-Instruct} \citep{bai2025qwen3vl}.
\paragraph{Software and search space.} PyTorch 2.6, transformers 4.57, scikit-learn 1.9.1 on the cluster and 1.8.0 locally, PyAV 13 for decoding; PyTorch 2.11 (CUDA 12.8) and transformers 4.57.6 on the eight-card RTX~5090 machines and PyTorch 2.14 on the single-card ones. We varied the predictor learning rate over $\{5\times10^{-5}, 10^{-4}\}$, the encoder learning rate over $\{3\times10^{-5}, 10^{-4}\}$, the mask weights over the four mixtures listed above, and the real-hand batch fraction over $\{0.25, 0.35\}$; the reported configurations are the first that trained stably, not the best of a sweep.

\paragraph{Compute.} The main experiments ran on NVIDIA H100 and H200 GPUs: continued pretraining takes 3.5\,h on four GPUs per run, the real-hand curriculum 3\,h on one, the container curriculum and each probe run under one GPU-hour, roughly 150 GPU-hours in total. The controls of Appendix~\ref{app:review} and the domain-matched curricula of Table~\ref{tab:intphys} ran on RTX~5090 GPUs, first on two eight-card machines shared with other jobs, on which the released ViT-L, ViT-H and ViT-g reproduce the cluster's 61.4, 84.2 and 54.2\%, then on four single-card machines, which evaluated the three-seed ViT-H curricula; about 60 GPU-hours in all. Curricula there used 12 clips per step on a dedicated card or four clips with three accumulation steps on a shared one. By block the released ViT-H reads O1 object permanence 80.0, O2 shape constancy 81.7 and O3 spatio-temporal continuity 90.8\%.
\paragraph{Linear probes.} Logistic regression (inverse regularisation strength 0.5) on token features with grouped cross-validation, events assigned to five folds by their sorted index. We fixed the fold assignment because the default grouped splitter changed between scikit-learn releases and moved the closed-hand accuracy by up to 0.16 on identical data.

\section{Additional Controls}
\label{app:controls}
\begin{figure}[h]
\centering
\includegraphics[width=\linewidth]{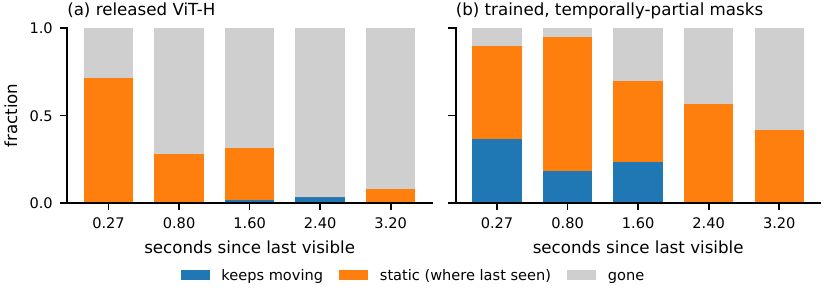}
\caption{Where the hidden moving object goes: nearest of three worlds over the hidden band. The released predictor says \emph{gone} within a second; the predictor trained with temporally-partial masks (v2) puts it where it was last seen from the second tubelet on; \emph{gone} is its most frequent answer only at 2.7--3.2\,s.}
\label{fig:threeway}
\end{figure}

The three further variants of the temporally-partial-mask predictor (a 32-frame window, a re-weighted loss, a 24-layer predictor) track the v2 curve within 0.12 at every point but one (the 32-frame variant reads 0.08 against 0.27 at 2.13\,s) and share its 1.1\,s carry-over (the 24-layer variant holds the \emph{static} world to the end of the window, 0.98 at 3.2\,s, where v2 reads mostly \emph{gone} at 2.7--3.2\,s); the never-visible control of v2 reads 0.18 then 0.00. Geometry-matched never-rendered controls (Appendix~\ref{app:review}) read 0.07 / 0.30 / 0.43 at 0\,s for the released ViT-L / ViT-H / ViT-g on a moving ball and 0.00--0.02 afterwards; on a stationary ball 0.42 / 0.67 / 0.80 at 0\,s, 0.10 / 0.23 / 0.02 at 0.27\,s and at most 0.07 afterwards. Never-visible controls for the moving-object probe: released ViT-L 0.12 then 0.00; tube-only continued pretraining 0.08 then 0.00; mixture 0.85, 0.25, 0.07, 0.00, 0.15, 0.08, 0.13, 0.42, 0.02, from which the 0.45 value of the mixture model at 1.87\,s in the moving curve is attributable to bias. Stationary never-visible controls: released ViT-L 0.43, 0.12, 0.05, 0.00; tube-only 0.58, 0.07, 0.03; mixture 0.98, 0.60, 0.12, 0.00, 0.10, 0.00, 0.13, 0.25, so the mixture model's stationary belief of 1.00 is above its control from 0.53\,s onwards. Never-visible controls for the container-curriculum and real-hand models in the reveal test: 0.32 at 0\,s and at most 0.02 afterwards. The second tube-only seed, the internet-style run and the three single-mask-kind predictors have no never-visible control of their own; their carry-over is read against the control of the closest sibling run (tube-only seed 1: 0.08 then 0.00; the no-container-mask curriculum's never-rendered control: 0.20--0.22 then at most 0.02). The mask mixture keeps a ball inside a moving and a stationary box at 1.00 at every hidden duration, against never-shown-box and exit controls (moving box) of 0.03 at 0.53\,s and 0.00 afterwards.

\section{Pixel-Reconstruction Readout}
\label{app:videomae}
VideoMAE (base and large, Kinetics-400; base, Something-Something~v2) with 16 frames at 224$\times$224, tubelet 2, 1{,}568 tokens. The released per-patch normalised MSE target makes any flat object match a zero-variance patch, so the never-visible control for an untextured 32\,px ball reads 0.73--1.00 over the first six tubelets; the checker-textured ball removes this and is used throughout. Pooled $\pres$ over the first six tubelets: large 0.35, 0.27, 0.03, 0.00, 0.00, 0.00; base 0.13, 0.07, 0.02, 0.05, 0.05, 0.03; base SSv2 0.13, 0.03, 0.00, 0.00, 0.00, 0.00; never-visible control (large) 0.10, 0.03, 0.00, 0.00, 0.00, 0.02. For the 32\,px ball alone VideoMAE-L reads 1.00, 0.93, 0.13 and then 0.00 (control 0.00), the 0.3\,s of Section~\ref{sec:paradigms}; the 16\,px ball reads 0.40, 0.13, 0.00 and smaller balls 0.00 throughout. V-JEPA~2 ViT-H on the textured ball: released 0.83, 0.05, 0.12, 0.00, 0.00, 0.00; with temporally-partial masks 1.00, 1.00, 0.93, 0.73, 0.78, 0.80.

\section{Autoregressive Readout}
\label{app:cosmos}
Cosmos-Predict1-4B is a 16-layer transformer of width 4{,}096 with 32 query and 8 key-value heads, RMSNorm on queries and keys before a 3D rotary embedding ($\theta=5\times10^{5}$, no YaRN for the 4B model), SwiGLU feed-forward blocks of width 14{,}336, a vocabulary of 64{,}000 video tokens and no special tokens. We re-implemented the forward pass from the released weights and checked each choice against the released source; per-token log-likelihoods of natural video tokens average $-5.5$ nats against $-11.1$ for a uniform distribution. Clips are 33 frames, the training extent of the model, tokenised by the released DV8$\times$16$\times$16 encoder into five latent chunks of 16$\times$16 tokens. The reveal is aligned to a chunk boundary. The tokenizer is documented as temporally causal, and the tokens before the event were identical between the two worlds in 29 of the 30 scenes of every condition; in the remaining scene, always the first one processed in a run, 52--470 prefix tokens differed (297--465 in the four container conditions), which points to a warm-up nondeterminism of the tokenizer call rather than to look-ahead. Both sequences therefore use the absent world's prefix tokens; repeating the comparison with the present world's prefix changes the mean evidence of each main condition by 0.05--0.16 nats and of each control by at most 0.47 nats. Each row of Table~\ref{tab:cosmos} pools three radii, two distractor levels and five seeds; resampling the 30 scenes gives 95\% intervals against the exit control of $-2.7$ [$-9.0$, 3.0] and $-10.3$ [$-17.4$, $-3.6$] nats for the moving box and $+20.7$ [14.5, 27.0] and $+18.7$ [13.3, 24.2] for the stationary one. The negative values for the moving box mean that a ball is rated less likely inside the box after it entered than after a ball left the frame; we have no explanation for the sign beyond the absence of a positive memory and do not interpret it further.

\section{Real-Hand Event Counts and Linear Probes}
\label{app:hands}
Events in the short-closure bin of Table~\ref{tab:hands}: EgoDex 8 of 19, EPIC-KITCHENS P27/P28 4 of 11, MagicCATs 39 of 81, MMC 11 of 19; a removed world exists for 3, 2, 24 and 9 of them. Closed-hand linear decodability over 0--0.3\,s (visible-coin control in parentheses): released ViT-H 0.56--0.59 (0.75--0.89); encoder trained, two worlds 0.73 (0.92); three worlds 0.88 (0.93); three worlds with containers 0.82 (0.89); clean recipe s1 / s2 on EgoDex 0.85 / 0.74 (0.94 / 0.95), on EPIC-KITCHENS P27/P28 0.75 / 0.78 (0.85 / 0.87), on MagicCATs 0.63 / 0.61 (0.80 / 0.78). The clean recipe (s2) reads 0.82 / 0.56 / 0.36 (coin-in / removed / never) on the 11 MMC events.

\section{Real-Magic Readouts}
\label{app:magic}
Cognitive science uses magic tricks as controlled violations of expectation \citep{kuhn2019experiencing}; MagicCATs \citep{ozono2021magiccats}, MMC \citep{meliss2024mmc} and the FunQA magic split \citep{xie2024funqa} provide clips with human surprise ratings, the ecological end of our evaluation.

No model we tested is surprised by a magic trick: across MMC, MagicCATs and the FunQA magic split, the prediction error of the released V-JEPA~2 at three scales under six readouts is indistinguishable from motion energy at the human-rated moments of surprise, and only 3 of the 99 rated MMC moments fall within one second of the re-opening of a detectable closing hand, so the probe rarely reads the interval a viewer reacts to (below).

Readouts on the released V-JEPA~2 at ViT-L, ViT-H and ViT-g: prediction-error time courses with context lengths of 2--10 tubelets, their motion-matched residuals, hand-region surprise, reveal surprise at the labelled reveal and the humour-versus-magic contrast on FunQA, and at ViT-H the belief probe over hand regions. None aligns with the human-rated moments beyond the alignment of motion energy; the belief probe over the hand regions of MMC clips gives a per-clip AUC of 0.49 for the released ViT-H and 0.44--0.49 for the trained checkpoints (16- and 32-frame windows; chance 0.50), and 11 of the 99 rated moments fall within two seconds of such a re-opening. Hidden intervals of the MMC and MagicCATs tricks, labelled by Qwen3-VL-8B-Instruct \citep{bai2025qwen3vl} at 2\,fps with greedy decoding and a fixed prompt (``This video shows a magic trick. Answer with two times in seconds, in the format HIDE=$\langle$s$\rangle$ REVEAL=$\langle$s$\rangle$. HIDE = the second at which the object (coin, ball, card, \ldots) last becomes hidden from view, for example when it is closed inside a hand, covered by a cup, or put behind something. If the object is never hidden, answer HIDE=none.'' and the matching definition of REVEAL), not checked against hand annotation, have medians of 1.5\,s (MMC, none shorter than 0.27\,s) and 3.0\,s (MagicCATs), and human surprise ratings in MagicCATs do not depend on them ($\rho=+0.05$ over 130 clips).

\section{Symmetric Readouts, Matched Controls and Curriculum Ablations}
\label{app:review}
Section~\ref{sec:encoder} reads the encoder with a trained linear probe and the predictor with the fixed nearest-reference rule of Section~\ref{sec:readout}. Three additions check that the gap is not made by that asymmetry, on the same 60 scenes per condition and the same hidden tokens. (a)~\emph{Encoder probe on $\hat z$}: the logistic probe fitted on the encoder's references for the hidden region ($z_{\text{present}}\to1$, $z_{\text{absent}}\to0$, one probe per time point, grouped-CV accuracy 0.98--1.00) is applied unchanged to the predictor's inference $\hat z$ for the same tokens. (b)~\emph{Probe on $\hat z$}: a probe fitted on $\hat z$ itself, main condition against a control, with the deterministic grouped five-fold split of Appendix~\ref{app:training}. (c)~\emph{Copy-last-view}: the encoder's tokens of the last visible slot at the same spatial patches, read with the paper's rule, which is what a model that only froze what it saw would score. The bias control here is geometry-matched: the object is never rendered and the mask is the main condition's (the never-visible control of Section~\ref{sec:controls} masks four more frames). The exit control of Section~\ref{sec:controls} is read on the same axes: a ball on the same row and at the same speed leaves the frame exactly when the mask starts (moving; no stationary version exists, hence the dashes) or when the box closes (container). Table~\ref{tab:review} gives ViT-H; ViT-L and ViT-g follow the same pattern (encoder probe on $\hat z$, moving: 0.95 and 1.00 at 0\,s, 0.00 from 0.27\,s on; stationary: 1.00 at 0\,s, then 0.00--0.28; $L_1$-rule never-rendered control 0.07 and 0.43 at 0\,s for a moving and 0.42 and 0.80 for a stationary ball, 0.00--0.10 afterwards; probe on $\hat z$ against the exit control 0.96--1.00 for a moving ball at both scales and, in container mode, 0.93--0.99 for ViT-L and 0.97--1.00 for ViT-g).

\begin{table}[h]
\centering\scriptsize
\caption{Symmetric readouts of the released ViT-H predictor's inference. Fraction of the 60 scenes read as \emph{present} (chance 0.5 for the probe on $\hat z$, which reports held-out accuracy of main against a control). Moving and stationary: seconds since the object was last visible; container: seconds hidden before the box opens, read at the opening slot. Copy-last-view: the encoder's tokens of the last visible slot, read with the $L_1$ rule; --: no stationary exit control.}
\label{tab:review}
\setlength{\tabcolsep}{2.5pt}
\begin{tabular}{llccccc@{\hspace{8pt}}cccc}
\toprule
 & & \multicolumn{5}{c}{moving / stationary, time since last visible (s)} & \multicolumn{4}{c}{container, time hidden (s)} \\
readout & condition & 0 & 0.27 & 0.53 & 1.07 & 2.93 & 0.53 & 1.07 & 2.00 & 2.93 \\
\midrule
$L_1$ rule (paper) & main & 0.92 / 1.00 & 0.00 / 0.83 & 0.18 / 0.63 & 0.03 / 0.52 & 0.00 / 0.25 & 0.05 & 0.02 & 0.00 & 0.00 \\
 & never rendered & 0.30 / 0.67 & 0.00 / 0.23 & 0.02 / 0.07 & 0.00 / 0.00 & 0.00 / 0.02 & 0.00 & 0.00 & 0.00 & 0.00 \\
 & exit & 0.30 / -- & 0.00 / -- & 0.02 / -- & 0.00 / -- & 0.00 / -- & 0.00 & 0.00 & 0.00 & 0.00 \\
copy-last-view & main & 0.98 / 0.92 & 1.00 / 0.80 & 0.92 / 0.98 & 0.58 / 0.98 & 0.08 / 1.00 & 0.37 & 0.35 & 0.30 & 0.28 \\
encoder probe on $\hat z$ & main & 0.68 / 1.00 & 0.22 / 0.50 & 0.08 / 0.02 & 0.00 / 0.58 & 0.00 / 0.18 & 0.02 & 0.00 & 0.00 & 0.00 \\
 & never rendered & 0.00 / 0.97 & 0.00 / 0.00 & 0.00 / 0.00 & 0.00 / 0.00 & 0.00 / 0.00 & 0.00 & 0.00 & 0.00 & 0.00 \\
 & exit & 0.00 / -- & 0.00 / -- & 0.00 / -- & 0.00 / -- & 0.00 / -- & 0.00 & 0.00 & 0.00 & 0.00 \\
probe on $\hat z$ & main vs. never rendered & 1.00 / 1.00 & 1.00 / 1.00 & 1.00 / 1.00 & 1.00 / 1.00 & 1.00 / 1.00 & 1.00 & 1.00 & 1.00 & 0.99 \\
 & main vs. exit & 1.00 / -- & 1.00 / -- & 1.00 / -- & 1.00 / -- & 1.00 / -- & 1.00 & 1.00 & 1.00 & 0.99 \\
\bottomrule
\end{tabular}
\end{table}

Under the encoder's own probe the predictor's output is read as containing the object in 68--100\% of scenes at the first hidden step and in at most 20\% (moving) from 0.53\,s on, against 0\% for the matched control (moving; the stationary control reads 97\% at 0\,s and 0\% afterwards); for a stationary object the same probe fires in 0--87\% of scenes from 0.27\,s on, and inside a closed box in 2\% at 0.53\,s and in none afterwards. The copy baseline keeps a stationary object at 0.80--1.00 throughout and a moving object's last position for one to two seconds (0.58 at 1.07\,s, 0.62 at 1.87\,s), so the released predictor scores below a model that merely froze its last view. A probe fitted on $\hat z$ separates the main condition from either control at 0.99--1.00 at every delay: the predictor's output is not blind to what the context showed, and what it lacks is the presence feature that the encoder's probe reads, on the axis that separates the two worlds.

\paragraph{Object size.} The loss does not depend on the size of the ball: from 0.53\,s on, the released ViT-H reads a moving ball as present in 3, 3, 1 and 2\% of scenes for radii of 4, 8, 16 and 32\,px, a stationary one in 39, 35, 40 and 51\% (never-rendered controls at most 7\%), and a ball in a closed box, at the opening slot, in 5, 2, 0 and 0\%.
\paragraph{Projection coefficient.} The signed distance $e$ of Equation~\ref{eq:evidence} is compressed by prediction error in every direction, so we also report the projection $\alpha=\langle\hat z-z_{\text{absent}},\,z_{\text{present}}-z_{\text{absent}}\rangle/\lVert z_{\text{present}}-z_{\text{absent}}\rVert^2$ of the region-mean prediction onto the axis between the two worlds, which is unbiased by error orthogonal to that axis (1 is the present world, 0 the absent one; the matched control sets the floor). For the released ViT-H and a moving ball, $\alpha$ reads 0.43, 0.38, 0.42 and 0.33 at 0.53, 1.07, 1.87 and 2.93\,s (0.48 at 0.27\,s) against 0.14, 0.12, 0.17 and 0.13 for the never-rendered control (0.14 at 0.27\,s) and 0.87, 0.70, 0.72 and 0.53 for a copy of the last visible tokens; for a stationary ball 0.46, 0.51, 0.57 and 0.44 against 0.17, 0.19, 0.23 and 0.16, with the copy at 0.93, 1.04, 0.79 and 0.91; at the opening slot of the visible-container test 0.35, 0.29, 0.22 and 0.16 after 0.53, 1.07, 2.00 and 2.93\,s hidden, against 0.23, 0.19, 0.15 and 0.12 for the never-shown box, 0.26, 0.21, 0.17 and 0.12 for the exit control and 0.52, 0.48, 0.46 and 0.41 for a copy of the closed box's tokens (paired $p<10^{-11}$ against the control at every point). The exit control reads 0.15--0.19 at the same four time points for a moving ball, within 0.03 of the never-rendered control, so the trace is specific to the ball's position and not to a ball having been in the scene. The prediction therefore carries a trace of the object: for a moving ball after the first tubelet and for a closed box it stays below the midpoint, which is why most decisions and the encoder's own probe read \emph{absent}, and for a stationary ball it hovers at the midpoint (0.44--0.57), as its decisions split. For a moving or stationary ball it lies 37--60\% as far above the control as the copy does, and inside a closed box 41\% after 0.53\,s and 14\% after 2.93\,s. ViT-L and ViT-g read the same way, their traces at 19--36\% and 36--59\% of the copy's for a moving ball, 31--37\% and 24--34\% for a stationary one and 30--37\% and 21--45\% inside a box. Training raises the trace: the three domain-matched curricula below (seed 2) reach 46--97\% of the copy on a moving ball, 64--87\% on a stationary one and 26--70\% inside a box, the moving curriculum highest on a moving ball, as the decisions rank it, and the static one highest on a stationary ball, where the decisions tie.
\paragraph{Pretraining-consistent query.} V-JEPA~2 is trained with masks that span every frame, whereas the probe hides the band only from the event on. To rule out that the released predictor fails only because it was never queried this way, the block ahead of the ball is masked in every frame of the window (a tube in V-JEPA's sense, starting where the ball enters it) and the ball is visible outside the block for 8 or 16 frames, 1.1 or 2.1\,s, before it enters; present and absent worlds differ only inside the block after entry, and the never-rendered control draws no ball under the same tube. For the released ViT-H, $\pres$ over the slots after entry reads 0.92, 0.10, 0.08, 0.07, 0.07, 0.03 with the 1.1\,s approach and 1.00, 0.52, 0.30, 0.22, 0.13, 0.03 with the 2.1\,s approach, against 0.47 then 0.00 for the control with either approach; ViT-L reads 0.37, 0.07, 0.00 and 0.85, 0.08, 0.02, 0.20, 0.00 against 0.32 then 0.00, and ViT-g 0.98, 0.42, 0.05, 0.05 and 1.00, 0.33, 0.20, 0.23, 0.08 against 0.52 and 0.68 then at most 0.02. The signed distance $e$ of the main condition stays above the control's at every slot (ViT-H, $-0.005$ against $-0.090$ at 0.27\,s). A longer visible approach buys ViT-H and ViT-L one further tubelet and ViT-g none, and the object is gone from their prediction within 0.8\,s under either geometry as long as the rest of the frame stays visible. Pretraining also hides about 90\% of the frame; hiding 90\% of the context outside the ball's approach as well (a random subset of tokens per scene, the approach band and the block excepted; 2.1\,s approach) leaves ViT-L and ViT-g where they were (0.77, 0.12, 0.18, 0.23 and 0.97, 0.18, 0.03, 0.02, against controls of 0.20 and 0.62 then at most 0.07) but lets ViT-H keep the ball for four tubelets after entry: 1.00, 0.87, 0.95, 0.87, 0.90, 0.27, 0.42, 0.20 against a control of 0.70 then 0.02 ($e$ positive through 1.07\,s; the control's $e$ negative from 0.27\,s on). With few same-instant neighbours to fill from, the ViT-H predictor uses the ball's past; ViT-L and ViT-g do not. The no-container-mask curriculum of Section~\ref{sec:install}, queried through the same tube after a 1.1\,s approach, keeps the ball at 1.00, 1.00, 0.98, 1.00, 0.98, 0.83, 0.85, 0.75 (ViT-H), 0.97, 1.00, 1.00, 0.98, 0.95, 0.95, 0.95, 0.97 (ViT-L) and 1.00, 0.98, 1.00, 0.98, 0.98, 0.97, 0.98, 0.95 (ViT-g), with controls of 0.00--0.50 at the first slot and at most 0.03 after; these curricula were themselves trained with tube masks on this renderer, so this shows that the query is answerable, not that the released model's failure is a query artefact (Appendix~\ref{app:scope}).
\paragraph{Frame rate.} The same scenes rendered at 15 and 30\,fps (128- and 256-frame clips, 64-frame windows, the same physical speed and the same 0.53\,s of visible object before the mask, 8 and 16 frames) give, for the released ViT-H and a moving ball, $\pres=1.00$, 0.40, 0.50, 0.13, 0.25, 0.03 at 0 to 0.67\,s in steps of 0.13\,s at 15\,fps, and 1.00, 0.93, 0.97, 0.65, 0.50, 0.05, 0.17, 0.00 at 0 to 0.47\,s in steps of 0.07\,s at 30\,fps, against matched never-rendered controls of 0.62 and 0.83 at 0\,s and at most 0.10 afterwards. For ViT-H the decision is last on the present side at 0\,s at 7.5\,fps and at 0.27\,s at 15 and 30\,fps (0.50 at both rates, after 0.40 at 0.13\,s at 15\,fps); ViT-L leaves it within one tubelet at every rate (0.47, 0.10 at 15\,fps and 1.00, 0.33, 0.03 at 30\,fps) and ViT-g after one or two (1.00, 0.52, 0.15 and 1.00, 0.48, 0.20; controls 0.08--0.92 at 0\,s and at most 0.07 afterwards), so the reach is a short time for ViT-H and one or two steps for ViT-L and ViT-g. A stationary ball reads 1.00, 0.87, 0.73, 0.65, 0.40 (ViT-H, 15\,fps) and 1.00, 0.92, 0.83, 0.48, 0.72 (30\,fps) over the first five slots and then fluctuates between 0.10 and 0.73, as it does between 0.17 and 0.63 at 7.5\,fps, with controls of 0.98 and 0.95 at 0\,s and at most 0.65 at the second slot; ViT-L reads 0.95, 0.63, 0.22 and 1.00, 0.92, 0.68, 0.77 and ViT-g 0.97, 0.27, 0.15 and 1.00, 0.57, 0.37.

\paragraph{Domain-matched curricula.} To split the IntPhys-2019 gain of Table~\ref{tab:intphys} into domain adaptation and persistence, the no-container-mask curriculum (same renderer, appearance randomisation, scenarios and mask mixture, 3{,}000 predictor-only steps of 12 clips from the released weights) is trained in two further forms. \emph{Nothing moves}: ball, box and distractors are frozen at their first-frame positions, so the scenes carry the renderer's appearance and objects that persist in place but no motion. \emph{Nothing persists}: ball, box and distractors are redrawn at independent random positions in every frame, so the appearance statistics are those of the curriculum and no object can be carried from one frame to the next; the background still persists. For ViT-H each curriculum was trained three times. With two context frames the released model reads 84.2\% (O1 80.0, O2 81.7, O3 90.8), the moving curriculum 93.9, 94.2 and 94.7\% (mean 94.3), the static one 93.6, 94.2 and 93.6\% (93.8) and the non-persistent one 88.9, 88.9 and 88.3\% (88.7); with the minimum over context lengths they read 93.1, 93.9 and 93.1\% (93.3), 90.0, 90.8 and 90.8\% (90.6) and 90.3, 91.7 and 91.1\% (91.0). Resampling the 90 scenes (each holds four pairs that share videos; 10{,}000 resamples; seed-averaged per scene) gives, with two context frames, 95\% intervals of $+4.5$ [0.6, 8.8] points for non-persistent minus released, $+5.1$ [1.9, 8.4] for static minus non-persistent and $+0.5$ [$-2.0$, 3.0] for moving minus static; under the minimum rule the static and non-persistent curricula lie within 0.5 points of each other and the moving one 2.3--2.8 points above both. ViT-g reads 54.2\%, 70.0\% and 77.5\% (released, non-persistent, moving; differences $+15.8$ [11.4, 20.6] and $+7.5$ [3.1, 12.2]) and ViT-L 61.4\%, 66.7\% and 67.8\% ($+5.3$ [1.4, 9.2] and $+1.1$ [$-1.4$, 3.6]), one seed each. For ViT-H under the two-frame rule, rigid motion adds nothing on top of persistence in place and domain adaptation accounts for about half of the gain (4.5 of 10.1 points); under the minimum rule persistence in place adds nothing on top of the domain and motion adds 2--3 points. Which habit the benchmark credits depends on the scoring rule under our implementation.

\paragraph{Belief probes on the curricula.} The three ViT-H curricula read the same on a hidden stationary ball, 0.88--1.00 at every step to 3.47\,s against matched never-rendered controls of at most 0.15 after the first step. They differ on a moving ball: at 0.27, 0.53, 1.07, 1.60 and 2.13\,s the moving curriculum reads 1.00, 0.98, 0.88, 0.78 and 0.38, the non-persistent one 0.95, 0.95, 0.58, 0.42 and 0.05, and the static one 0.68, 0.72, 0.18, 0.03 and 0.00. None keeps a ball in a closed box (0.00--0.30 against controls of 0.00--0.12). With two context frames the benchmark ranks the static curriculum with the moving one and above the non-persistent one; the probe ranks it last on a moving object and ties all three on a stationary one. What the non-persistent curriculum lacks for the benchmark is therefore not what the probe reads; one candidate is that its objects jump in every frame, which makes a jump in a test video less surprising.

\paragraph{Protocol.} Every IntPhys-2019 number in the main text uses one setting, two context frames and the surprise averaged over windows, chosen for no model, unless it is labelled as the minimum rule (Sections~\ref{sec:objective} and~\ref{sec:install}). \citet{garrido2025intuitive} report this fixed-context variant in their supplement but select the context length per property and, on IntPhys, take the minimum surprise over context lengths for each start frame. Table~\ref{tab:protocol} gives the three rules side by side. Every curriculum gain survives every rule, but the order among the domain-matched curricula and among the natural-video interventions changes with it: under the minimum rule the moving curriculum leads, the static and non-persistent ones tie, and the mask mixture overtakes the released ViT-L, which falls below chance. Our implementation of the minimum rule takes the minimum over context lengths 2--10 at one window size and frame rate, whereas \citet{garrido2025intuitive} also sweep window sizes and frame skips; it does not reproduce their published numbers (the released ViT-H reads 70.0\% against 87--98\% reported for V-JEPA models), so we use it only to compare our own checkpoints under a second rule. The cluster and the eight-card RTX~5090 machines give identical released baselines (61.4, 84.2 and 54.2\%); re-evaluating the same checkpoint on the single-card machines, with a newer PyTorch, moved accuracy by up to 1.1 points (bf16 numerics; the released ViT-L and ViT-g read 61.7 and 53.1\% there, 45.3 and 48.9\% under the minimum rule). Only the three-seed ViT-H rows of Tables~\ref{tab:intphys} and~\ref{tab:protocol} come from those machines; the released ViT-H was not re-evaluated there, so its comparison with those rows mixes machines. On IntPhys~2 (48-frame windows at 6\,fps, context lengths 12--42) the released ViT-H reads 52.8--54.0\% across the six context lengths and the container curriculum 51.2--54.2\%.

\begin{table}[h]
\centering\small
\caption{IntPhys-2019 pairwise accuracy (\%) under three aggregation rules: the fixed two-frame context of the main text, the best single context length in 2--10 frames chosen per model, and the minimum surprise over those context lengths per start frame (our implementation of the rule of \citealp{garrido2025intuitive}); average surprise throughout. $^*$Evaluated on the single-card RTX~5090 machines (Appendix~\ref{app:training}), mean of three seeds trained there; the original run is a separate seed trained and scored on the cluster; the two-seed rows list each seed.}
\label{tab:protocol}
\setlength{\tabcolsep}{2.5pt}\footnotesize
\begin{tabular}{llccc}
\toprule
 & & fixed $C=2$ & best fixed $C$ & min over $C$ \\
\midrule
ViT-H & released & 84.2 & 84.2 ($C=2$) & 70.0 \\
 & + container curriculum & 93.3 & 93.3 ($C=2$) & 91.4 \\
 & + no container masks (original run) & 93.9 & 94.2 ($C=6$) & 93.6 \\
 & \quad moving / static / non-persist.$^*$ (3 seeds) & 94.3 / 93.8 / 88.7 & 94.7 / 93.9 / 90.6 & 93.3 / 90.6 / 91.0 \\
 & + real-hand clean recipe (2 seeds) & 27.2 / 30.6 & 42.8 / 47.8 ($C=10$) & 38.6 / 43.9 \\
\midrule
ViT-L & released & 61.4 & 61.4 ($C=2$) & 45.0 \\
 & + container curriculum & 64.4 & 69.4 ($C=8$) & 69.4 \\
 & + tube masks, 16k (2 seeds; internet) & 62.8 / 64.2 / 62.5 & same ($C=2$) & 59.2 / 61.9 / 59.2 \\
 & + mask mixture, 16k & 53.1 & 66.9 ($C=10$) & 63.1 \\
\midrule
ViT-g & released & 54.2 & 54.2 ($C=2$) & 48.9 \\
 & + container curriculum & 75.0 & 76.9 ($C=10$) & 75.0 \\
\bottomrule
\end{tabular}
\end{table}

\section{Scope, Limitations and What Would Change Our Conclusions}
\label{app:scope}
This appendix states, in one place, the boundary of every claim in the paper and the measurement that would move it.

\paragraph{What was probed.} The object of study is the masked-prediction predictor of the released V-JEPA~2 checkpoints, together with the pixel decoder of VideoMAE and the token predictor of Cosmos-Predict1-4B, all frozen. We did not probe V-JEPA~2-AC, the action-conditioned predictor trained on the same frozen encoder for planning \citep{assran2025vjepa2}; V-JEPA~2.1, whose dense loss also supervises visible tokens \citep{murlabadia2026vjepa21}; diffusion world models; VideoMAEv2; or any model with an explicit memory, such as WorldMem \citep{xiao2025worldmem}, state-space world models \citep{po2025longcontext} or slot models. ``Video world models'' in the title therefore means the three prediction paradigms tested, not the field; a positive control with built-in memory would calibrate the probe's ceiling and is the first addition we would make.

\paragraph{Where it was probed.} Every belief curve comes from one renderer: a single row, one speed, four radii, tiled backgrounds and rigid boxes, 60 scenes per time point. Natural video enters only through the container curriculum evaluated over natural backgrounds (0.85--0.97) and the composited coins on real hands, where the held-out sets hold 8 and 4 events (3 and 2 with a removed world) and the released encoder itself loses the coin once the hand closes (Section~\ref{sec:encoder}). We have not read the released models with a two-world test on real occlusion video such as the real test set of TCOW \citep{vanhoorick2023tcow} or the cups game of the Perception Test \citep{patraucean2023perception}; the hand-region readouts on magic clips of Appendix~\ref{app:magic} are not such a test, and the finding of Section~\ref{sec:curves} is a synthetic-scene finding until that is done.

\paragraph{Decision against trace.} ``Lost within 0.3\,s'', ``in part'' and ``not at all'' describe the two-alternative decision. The projection of Appendix~\ref{app:review} shows a trace at 14--60\% of the copy baseline that decreases slowly (for a moving ball, ViT-H reads 0.43 at 0.53\,s and 0.33 at 2.93\,s), specific to the ball's position and, for a moving ball after the first tubelet or a closed box, below the midpoint. Which of the two readouts ``is'' permanence is a choice; we report both and lead with the decision because it is what a downstream reader of the predictor's output obtains. A calibration on a predictor trained with a known posterior, where $\pres$, $e$ and $\alpha$ can be compared with the truth, would settle how each readout maps onto belief and is not in this paper. Carry-over durations are threshold crossings of a 60-scene proportion whose binomial half-width near 0.5 is 0.13, so they are resolved to about one tubelet, and we give no interval for them.

\paragraph{Query geometry.} The main mask hides the band only after the event, which V-JEPA~2 never trained on; the tube query of Appendix~\ref{app:review} hides a block in every frame but leaves about 90\% of the frame visible, whereas V-JEPA's multi-block masks hide about 90\% of it \citep{bardes2024vjepa}; the variant that also hides 90\% of the context outside the ball's approach is reported there, and it changes the answer for ViT-H (1.1\,s of carry-over) but not for ViT-L or ViT-g. The visible history before the mask is 0.53\,s in the main probe and 1.1--2.1\,s in the tube query; longer histories, other speeds, and the identity of the released encoder (context or EMA target, which we did not determine) remain to be varied or determined. For balls of 16\,px and more, the moving and static worlds overlap the readout region in the first two tubelets, so the three-way readout of Figure~\ref{fig:threeway} separates them only later.

\paragraph{Mechanism.} That tube masking lets the predictor fill a block from the same instant (Section~\ref{sec:reveal}) is a hypothesis consistent with the reveal test and the sparse-context query (ViT-H uses the past once its same-instant neighbours are hidden), not a demonstrated cause; the mask-kind control does not support it, since tube masks alone gave 1.6\,s of carry-over on synthetic scenes; continued tube-mask training under our recipe did produce carry-over (Section~\ref{sec:objective}), and which of clip length, masking ratio, EMA initialisation or clip count separates our recipe from the released one is unmeasured. ``What persists changes with the masks and the data'' is the observation that different mixtures and sources gave different curves, not an isolated cause.

\paragraph{Interventions.} The synthetic curricula train and test on the same renderer; that a 3{,}000-step curriculum reaches 1.00 on its own distribution is expected, and what it installs is a prior that a reveal of the empty band does not revise (Section~\ref{sec:reveal}). The non-persistent curriculum keeps a stationary ball at 0.88--1.00 and a moving one at 0.95 at 0.27 and 0.53\,s (0.58 at 1.07\,s), so on this renderer the probe cannot separate learned persistence from adaptation to the domain and a copying strategy; held-out renderers and real video are needed for that. No intervention was evaluated for what it costs the representation (attentive probes on Kinetics or Something-Something, action anticipation, planning), and the real-hand recipe's IntPhys-2019 accuracy of 27--31\% is evidence that some cost exists. ``Cheaply'' refers to the step count, not to a measured trade-off.

\paragraph{IntPhys-2019.} All comparisons use the 360 dev pairs; there is no held-out split. Our minimum-over-contexts rule does not reproduce published numbers (Appendix~\ref{app:review}), so the rule-dependence we report is a statement about our checkpoints under our two rules. Released ViT-L and ViT-g read 61.4\% and 54.2\%, far below ViT-H and for ViT-g close to chance, which we do not explain. IntPhys~2 shows no gain (51.2--54.2\% against 52.8--54.0\%). The probe is in-distribution for the curricula and the benchmark is not, which is one reason their rankings may differ.

\paragraph{Statistics.} $\pres$ values are proportions over 60 scenes (30 for Cosmos) with binomial intervals; comparisons of $e$ and $\alpha$ against a control use paired $t$-tests over the shared scenes at matched time points; IntPhys-2019 differences use a scene-level bootstrap. No correction is applied for the many time points and conditions compared, so the $p$-values describe the strength of a paired separation rather than confirm a pre-registered hypothesis, and the curves of the released models are deterministic given the scenes, so their seeds are the scene seeds, not model seeds. Single-run interventions are listed in Section~\ref{sec:train}.
\paragraph{Other paradigms.} Only VideoMAE v1 was tested, and Cosmos only at 256$\times$256, below its training resolution, with a re-implemented forward pass checked against natural-video log-likelihoods but not aligned token by token with the official code; the negative memory of the moving box is unexplained. These two readouts support the narrow statements made about them and no cross-paradigm generalisation beyond that.

\paragraph{Related and concurrent work not compared.} Layer-wise probing of physical variables in video encoders \citep{joseph2026interpreting,punzo2026probing}, the observation that video diffusion models hold physical information they do not render \citep{esmati2026invisible}, and benchmarks of unobserved-state tracking in world models \citep{lu2026persistent,shin2026track} ask neighbouring questions with other models and readouts; HERA \citep{yuanruyi2026hera} adds a memory adapter to a frozen V-JEPA~2 predictor. Object-level models that handle occlusion explicitly \citep{riochet2020occlusion} and physical-law evaluations of generative video models \citep{kang2025howfar,motamed2025physicsiq} are the natural comparisons for a broader version of Section~\ref{sec:paradigms}. \citet{garrido2025intuitive} ablate pretraining masks on IntPhys (their Figure~3B, robustness to the mask) and raise the hidden-object question in their appendix; Sections~\ref{sec:curves} and~\ref{sec:objective} are the corresponding measurements at the level of the predictor's state.

\paragraph{What would change the conclusions.} The 90\%-ratio tube query already confines the moving-object loss of Section~\ref{sec:curves} to dense contexts for ViT-H; a released predictor that keeps a hidden object on real occlusion video would confine it to synthetic scenes. A positive-control model that reads 1.00 on the probe would fix its ceiling. A recipe ablation from the released weights that reproduces the carry-over with one changed factor would turn ``changes with the masks and the data'' into a cause. A held-out renderer on which the container curriculum keeps 1.00 would turn ``installs a prior'' into ``installs containment''. An IntPhys-2019 implementation that reproduces the published numbers under the minimum rule, and a held-out split, would decide whether the rule-dependence of Section~\ref{sec:install} is a property of the benchmark or of our implementation.

\end{document}